\documentclass[%
  onecolumn
   , hidempi
]{mpi2015-cscpreprint}
\makeatletter
\DeclareOldFontCommand{\rm}{\normalfont\rmfamily}{\mathrm}
\DeclareOldFontCommand{\sf}{\normalfont\sffamily}{\mathsf}
\DeclareOldFontCommand{\tt}{\normalfont\ttfamily}{\mathtt}
\DeclareOldFontCommand{\bf}{\normalfont\bfseries}{\mathbf}
\DeclareOldFontCommand{\it}{\normalfont\itshape}{\mathit}
\DeclareOldFontCommand{\sl}{\normalfont\slshape}{\@nomath\sl}
\DeclareOldFontCommand{\sc}{\normalfont\scshape}{\@nomath\sc}
\makeatother

\usepackage[american]{babel}

\usepackage{graphicx}
\usepackage{amssymb}
\usepackage{amsthm}
\usepackage[mathscr]{eucal}

\numberwithin{equation}{section}
\numberwithin{figure}{section}
\numberwithin{table}{section}

\usepackage[software,hardware]{mymacros}
\usepackage{import}
\usepackage{xifthen}
\usepackage{pdfpages}
\usepackage{transparent}
\usepackage{cleveref}
\usepackage{sidecap}    
\usepackage{floatrow}
\usepackage{todonotes}
\usepackage{algorithm}
\usepackage[noend]{algpseudocode}

\makeatletter
\def\algbackskip{\hskip-\ALG@thistlm}
\makeatother

\usepackage{caption}
\usepackage{subcaption}
\usepackage{arydshln}

\definecolor{lightgray}{gray}{0.9}
\usepackage{color}
\definecolor{bluegreen}{rgb}{0.0, 0.87, 0.87}
\usetikzlibrary{fadings,shapes.arrows,shadows}   
\usetikzlibrary{shapes.multipart}

\tikzfading[name=arrowfading, top color=transparent!0, bottom color=transparent!95]
\tikzset{arrowfill/.style={top color= black!20, bottom color=green!50!black, general shadow={fill=black, shadow yshift=-0.8ex, path fading=arrowfading}}}
\tikzset{arrowstyle/.style={draw=black,arrowfill, single arrow,minimum height=#1, single arrow,
		single arrow head extend=.2cm,}}

\newcommand{\rk}{{\sffamily{RK4}}}

\newcommand{\dt}{\text{\sffamily{dt}}}

\newtheorem{remark}{Remark}

\begin{document}
  

\title{Learning Low-Dimensional Quadratic-Embeddings of High-Fidelity Nonlinear Dynamics using Deep Learning}
  
\author[$\ast$]{Pawan Goyal}
\affil[$\ast$]{Max Planck Institute for Dynamics of Complex Technical Systems, 39106 Magdeburg, Germany.\authorcr
  \email{goyalp@mpi-magdeburg.mpg.de}, \orcid{0000-0003-3072-7780}
}
  
\author[$\dagger\ddagger$]{Peter Benner}
\affil[$\dagger$]{Max Planck Institute for Dynamics of Complex Technical Systems, 39106 Magdeburg, Germany.\authorcr
  \email{benner@mpi-magdeburg.mpg.de}, \orcid{0000-0003-3362-4103}
}
\affil[$\ddagger$]{Otto von Guericke University,  Universit\"atsplatz 2, 39106 Magdeburg, Germany\authorcr
  \email{peter.benner@ovgu.de} 
  \vspace{-0.5cm}
}
  
\shorttitle{Low-Dimensional Quadratic-Embeddings for Nonlinear Dynamics}
\shortauthor{P. Goyal, P. Benner}
\shortdate{}
  
\keywords{Machine learning, deep learning, autoencoders, high-fidelity dynamical systems, low-dimensional embedding, quadratic models.}

  
\abstract{%
Learning dynamical models from data plays a vital role in engineering design, optimization, and predictions. Building models describing dynamics of complex processes (e.g., weather dynamics, or reactive flows) using empirical knowledge or first principles are onerous or infeasible.  Moreover, these models are high-dimensional but spatially correlated. 
It is, however, observed that the dynamics of high-fidelity models often evolve in low-dimensional manifolds. Furthermore, it is also known that for sufficiently smooth vector fields defining the nonlinear dynamics,  a quadratic model can describe it accurately in an appropriate coordinate system, conferring to the McCormick relaxation idea in nonconvex optimization.
Here, we aim at finding a low-dimensional embedding of high-fidelity dynamical data, ensuring a simple quadratic model to explain its dynamics.  To that aim, this work leverages deep learning to identify low-dimensional quadratic embeddings for high-fidelity dynamical systems. Precisely, we identify the embedding of data using an autoencoder to have the desired property of the embedding. We also embed a Runge-Kutta method to avoid the time-derivative computations, which is often a challenge. We illustrate the ability of the approach by a couple of examples, arising in describing flow dynamics and the oscillatory tubular reactor model. 
}

\novelty{
	\begin{enumerate}
		\item Learning low-dimensional quadratic-embeddings for high-dimensional dynamical data using deep learning.
		\item Employing the power of neural networks, precisely, autoencoders, to identify low-dimensional embeddings such that a simple quadratic model can explain the dynamics of the embedding.
		\item Blending the fourth-order Runge-Kutta scheme to avoid time-derivative computations.
		\item By virtue of nonlinear projection, which is an intrinsic interpretation of autoencoders, low-dimensional representation can be found for data with a slowly decaying \emph{Kolmogorov $n$-width}.
		\item Showcasing the performance of the approach over finding the best possible quadratic models for low-dimensional embeddings using linear projection (via POD).
		\item Simple low-dimensional quadratic models should facilitate control prediction and optimization of high-fidelity nonlinear dynamical processes.
\end{enumerate}
} 
\maketitle

\section{Introduction}\label{sec:introduction}
Inferring mathematical models describing the underlying dynamical behavior is essential in building infrastructure to take technology forward. These models allow us to understand the underlying dynamics and to perform engineering studies, control, and predictions. Traditionally, such modeling is done based on the first principles (e.g., conservation laws, gravitational laws) and empirical knowledge by experts. However, for complex phenomenons and modern engineering tools (e.g., advanced robotic dynamics, climate dynamics), the first principle and empirical knowledge are not fully available to obtain dynamics models accurately. However, with advancements in measurement technology, data related to physical processes can be obtained, which can support in uncovering the underlying dynamics. The paradigm of modeling using solely data  has got a lot of attention in last couple of decades \cite{bongard2007automated,yao2007modeling,schmidt2009distilling,rowley2009spectral,schmid2010dynamic,morPehW16,morBenGW15,brunton2016discovering,williams2015data,morKutBBetal16,rudy2017data,pathak2018model,morBenGKetal20}. A key towards learning dynamics models is to obtain a parsimonious model which gives a good balanced between data fitting and model complexity. Though dictionary-based sparsity methods are prone to parsimonious models by enforcing sparsity, in our work, we impose parsimony intending to determine obtaining low-dimensional models with simple non-linearity, namely, quadratic-type.  The motivation behind this is that high-dimensional dynamical systems involve in low-dimensional manifolds \cite{morBenMS05,morPatR07,morSchVR08,morQuaMN16}; and, a nonlinear system with sufficient continuity can be written as a quadratic nonlinear system \cite{morGu11,qian2020lift}. It is related to  McCormick relaxation~\cite{mccormick1976computability}.

Neural networks based learning of dynamical systems from time-series data has surprisingly rich history \cite{SuyVdM96,NarP90,lusch2018deep,mardt2018vampnets,vlachas2018data,yeung2019learning,takeishi2017learning,champion2019data,morGoyB21}. Despite their success, interpretability and generalizability remains primary concerns. Parsimonious models, having the simplest model describing dynamics goes in the opposite direction of neural networks as neural networks are typically high-dimensional or highly parameterized.  Towards identifying interpretable and generalizable models, symbolic regression-based approaches have shown their potential \cite{bongard2007automated,yao2007modeling,schmidt2009distilling}. A seminal approach, the dictionary-based sparse regression also emerged as a powerful method \cite{schaeffer2013sparse,brunton2016discovering}. It relies on constructing a large feature dictionary followed by identifying the correct few features from the dictionary that describe the system dynamics. The approach is computationally efficient, but the success of the method highly relies on the quality of the dictionary. The work \cite{champion2019data} takes it one step further and aims to identify the transformation of coordinate such that the dynamics can be given sparsely in the constructed dictionary. 

Furthermore, several complex dynamical phenomena are high-dimensional and highly nonlinear. It makes analysis and control of those models \emph{almost} infeasible.  However, a redeeming of these high-dimensional systems is that their dynamics lie in a low-dimensional manifold. Principle component analysis (\textsf{PCA}) is a widely used tool to obtain low-dimensional representation using linear projection. Dynamic mode decomposition \cite{schmid2010dynamic,morKutBBetal16,morBenHM18} aims at identifying a linear dynamical models in that low-dimensional, but often it is not sufficient to capture complex dynamics completely. 
Another popular method to obtain nonlinear dynamical models in the low-dimensional data  is operator inference (\textsf{OpInf}) \cite{morPehW16,morBenGKetal20}. The \textsf{OpInf} is also combined with hand-designed features such that the dynamics in the low-dimensional dynamics can be given by a quadratic model \cite{qian2020lift}. The \textsf{OpInf} is not fully data-driven as they require the form of governing equations that may not be available. In any case, these methods build on the linear projection of the high-dimensional data. However, often, a linear projection does not give a good low-dimensional representation (e.g., advection-dominant problems) due to slow decay of Kolmogorov $n$-width. A nonlinear extension of \textsf{PCA}, the so-called autoencoder, is powered by neural networks. The autoencoder has been widely used to obtain low-dimensional representation, see, e.g., \cite{lee2020model,gonzalez2018deep}, but they are often decoupled with the course of learning dynamical models in the low-dimensional.  We firmly emphasize that there is not enough evidence that a simple model can explain the dynamics of the lowest possible dimensional embedding of data. Towards incorporating dynamics with autoencoder,  the work \cite{lusch2018deep} utilized the Koopman theory \cite{koopman1931hamiltonian} to identify the low-dimensional representation such that the dynamics are almost linear in the representation. Although analysis and engineering studies have become easier, linear models are not expressive enough to capture complex dynamics completely.  Furthermore, autoencoders are combined with sparse regression in \cite{champion2019data}, where the idea is to find the projection of the data such that dynamics of the projected data can be given by selecting few features from the dictionary. Despite being a promising approach, the dictionary-based approach quickly becomes intractable when the number of variables becomes even slightly larger (e.g., for $10$ variables and degree $3$ polynomial features, total features would be $286$). Additionally, the approach \cite{champion2019data} requires the derivative information, which is often a challenging task. 

In this work, we present a framework for learning a low-dimensional embedding of data such that the dynamics in the low-dimensional embedding can be given a simple quadratic model. It combines two main philosophies: (a) dynamics often lies in a low-dimensional manifold \cite{morBenMS05,morSchVR08,morQuaMN16},  and (b) continuous-enough dynamical systems can be written a quadratic model \cite{mccormick1976computability,morGu11}. We leverage the autoencoder neural network framework to simultaneously identify the desired low-dimensional embedding and the underlying quadratic model. It advocates a good balance between the low-dimensional embedding of data and the complexity of models, describing the dynamics of the embedding. Furthermore, to avoid time-derivative computations, we fuse a numerical integration method, the so-called \emph{Runge-Kutta scheme} to learn continuous quadratic models.  We showcase the approach with two examples: Burgers' equations describing flow dynamics and the tubular reactor model. These results expose the ability of our approach to obtain low-dimensional parsimonious models that contain only quadratic non-linearity.  

The remainder of the paper is organized as follows. In \Cref{sec:lifting}, we explain the writing of nonlinear dynamical systems as equivalent quadratic systems by an appropriate transformation of coordinates--also known as lifting. \Cref{sec:quad_encoding} contains our core contributions. There, we present autoencoder network architecture to identity low-dimensional embedding such that a simple quadratic model can describe its dynamics. We also discuss how a Runge-Kutta scheme can be fused in the course of learning encoding and decoding to compute continuous quadratic models to describe the embedding dynamics.  Thus, we can avoid any computation related to time-derivatives. We demonstrate the approach by means of two examples, namely, reactor tabular model and 2D Burgers' equation in \Cref{sec:numerics}. We summarize our work and provide discussions with future directions in  \Cref{sec:conclusion}.

\section{Quadratic Modeling of Nonlinear Systems}\label{sec:lifting}
We here briefly overview quadratic modeling of nonlinear systems and demonstrate that the sufficiently continuous nonlinear systems can be rewritten as quadratic systems. Such an approach is often employ to simplify nonlinear optimization problems \cite{mccormick1976computability} or model reduction for nonlinear systems \cite{morGu11,morBenB15}. Let us consider a nonlinear system as follows:
\begin{equation}\label{eq:NL_eqn}
	\dot \bx(t) = \mathbf f(\bx),
\end{equation}
where $\bx \in \Rn$ and the function $\mathbf f(\cdot): \Rn \rightarrow \Rn$ with $\mathbf f(\cdot)$ being sufficiently continuous.  Then, there exit a lifting mapping $\cL: \Rn \rightarrow \Rm$, and its inverse mapping $\cL^\sharp: \Rm \rightarrow \Rn$, resulting into
\begin{equation}\label{eq:NL_eqn_quad}
	\dot \bz(t) = \cA(\bz(t)) + \cH(\bz(t)),
\end{equation}
where $\bz(t) =  \cL(\bx)$, and  $\cL^\sharp\left(\cL(\bx) \right)= \bx$. Moreover, $\cA(\cdot)$ and $\cH(\cdot)$ are linear and quadratic operators, i.e., $ \cA(\bz(t)) = \bA \bz $ and $\cH(\bz(t)) = \bH \left(\bz(t)\otimes \bz(t)\right)$ with $`\otimes'$ denoting the Kronecker product \cite{kolda2009tensor}. We illustrate the whole philosophy with a simple example nonlinear system, describing dynamics of a simple pendulum:
\begin{equation}
	\begin{bmatrix} \dot x_1 \\ \dot x_2 \end{bmatrix} = \begin{bmatrix} -\sin(x_2)\\ x_1 \end{bmatrix}. 
\end{equation}
For the above system, we define transformed or lifted  coordinates and inverse transformation as follows:
\begin{equation}
	\cL:  \begin{bmatrix}  x_1 \\  x_2 	\end{bmatrix} \mapsto \begin{bmatrix}  x_1 \\  x_2 \\ \sin(x_2) \\  \cos(x_2)	\end{bmatrix} =: \begin{bmatrix}  z_1 \\ z_2 \\ z_3 \\  z_4	\end{bmatrix}, \qquad 	\cL^\sharp:  \begin{bmatrix}  z_1 \\  z_2 \\ z_3 \\ z_4 	\end{bmatrix} \mapsto \begin{bmatrix}  z_1 \\  z_2 	\end{bmatrix} \equiv \begin{bmatrix}  x_1 \\  x_2 	\end{bmatrix}.
\end{equation} 
Consequently, we can write dynamics in the variables $z_i$'s as a quadratic system:
\begin{equation}
	\dfrac{d}{dt}\begin{bmatrix}  z_1 \\ z_2 \\ z_3 \\  z_4	\end{bmatrix} = \begin{bmatrix}  -z_3 \\ z_1 \\ z_1z_4 \\  -z_1z_3	\end{bmatrix}.
\end{equation}
The motivation for having quadratic models is that they ease analysis and engineering studies compared to complex nonlinear systems.  For more details on lifting and writing nonlinear systems as quadratic systems, we refer to \cite{mccormick1976computability,morGu11,morBenB15,qian2020lift}. However, we like to highlight that the lifting transformation to rewrite the dynamics as quadratic systems is not unique, and the inverse transformation is not either. Moreover, we need a nonlinear analytical system to write a lifted quadratic system, which is visibly not available where our aim itself is to find a model from data. 
\section{Autoencoders and Quadratic Embeddings for Learning Nonlinear Dynamics}\label{sec:quad_encoding}
As noted, nonlinear systems can often be written as quadratic systems. However, it requires the information of lifting transformation and its inverse transformation. In other words, it demands a coordinate change such that a simple quadratic model can describe dynamics in the transformed coordinate. Limiting ourselves to such a model, we obtain a parsimonious model prone to interpretability and generalizability. Also, we focus on exploiting the fact that the dynamics of high-fidelity models often lie in a low-dimensional manifold.  Thus, our work leverages the impressive approximation capabilities of deep neural networks to obtain a low-dimensional embedding.
Many low-dimensional embedding may exist, but it is not necessary that a quadratic model can explain the dynamics in the obtained low-dimensional embedding. Hence, our primary goal is to discover a particular low-dimensional embedding so that a simple quadratic model can explain the dynamics of the embedding.

Consider a high-fidelity nonlinear dynamical system of form \eqref{eq:NL_eqn}. We seek to identify a low-dimensional embedding or coordinate $\bz(t) \in \R^{\hn}$ such that $\bx(t) \approx \Phi(\bz(t))$, where $\Phi: \R^{\hn} \rightarrow \Rn$ is a nonlinear function. Furthermore, $\bz(t)$ satisfies
\begin{equation}\label{eq:QuadModel_z}
	\dfrac{d}{dt}\bz(t) = \bA\bz(t) + \bH \left(\bz(t)\otimes \bz(t)\right) + \bb,
\end{equation}
where $\bA \in \R^{\hn\times \hn}, \bH \in \R^{\hn\times \hn^2}$, and $\bb \in \R^{\hn}$. Additionally, we require an inverse mapping, reconstructing $\bz$ from $\bx$, and we do this by another function $\Psi:\R^{n} \rightarrow \R^{\hn}$, i.e., $\bz(t) \approx \Psi(\bx(t))$. Consequently, we have an intrinsic autoencoder structure with a constraint on the low-dimensional embedding.  We aim to learn both functions, namely $\Phi$ and $\Psi$, using neural networks. Thus, it can be seen as a nonlinear projection of the high-dimensional data which can interpreted a non-linear generalization of  \textsf{PCA}  \cite{baldi1989neural}. As a result, our aim at find projections or functions $\Phi$ and $\Psi$, as well as, the matrices $\bA,\bH$ and $\bb$, describing the dynamics of the low-dimensional embedding. 

\begin{figure}[!tb]
	\tikzset{arrow/.style={-stealth, thick, draw=gray!80!black}}
	
	\begin{tikzpicture}
		
		\node[fill=blue!20, minimum width=0.5cm, minimum height=3.5cm] (X) at (0,0) {$\mathbf x$};
		
		\draw[fill=purple!20] ([xshift=0.5cm]X.north east) -- ([xshift=2.5cm,yshift=0.5cm]X.east) -- ([xshift=2.5cm,yshift=-0.5cm]X.east) -- ([xshift=0.5cm]X.south east) -- cycle; 
		\node at (1.75cm,0) {\textsc{Encoder}};
		
		\node[fill=red!20, minimum width=0.5cm, minimum height=1.0cm] (Z) at (3.5cm,0) {$\mathbf z$};
		
		\draw[fill=purple!20] ([xshift=0.5cm]Z.north east) -- ([xshift=2.5cm,yshift=1.25cm]Z.north east) -- ([xshift=2.5cm,yshift=-1.25cm]Z.south east) -- ([xshift=0.5cm]Z.south east) -- cycle;
		\node at (5.25cm,0) {\textsc{Decoder}};
		
		\node[fill=blue!20, minimum width=0.5cm, minimum height=3.5cm] (Xp) at (7,0) {$\mathbf{{x}}$};
		
		\draw[arrow] (X.east) -- ([xshift=0.5cm]X.east);
		\draw[arrow] ([xshift=-0.5cm]Z.west) -- (Z.west);
		\draw[arrow] (Z.east) -- ([xshift=0.5cm]Z.east);
		\draw[arrow] ([xshift=-0.5cm]Xp.west) -- (Xp.west);
		
		\draw[arrow] (Z.south) -- ([yshift=-0.75cm]Z.south);
		
		\node[fill=cyan!20, minimum width=0.5cm, minimum height=.5cm] (Xp) at ([yshift=-1.2cm]Z.south) {$\dot{\mathbf{z}} = \bA\bz + \bH\left(\bz\otimes \bz \right) + \bb$};
		
	\end{tikzpicture}
	\caption{The figure depicts a schematic of the approach. It involve an encoder, mapping the high-dimensional state $\bx$ to a dimensional state $\bz$, and a decoder, mapping back  $\bz$ to $\bx$. We enforce a constraint of the low-dimensional variable $\bz$---that is, a simple quadratic model can explain the dynamics of the variable $\bz$.}
\end{figure}
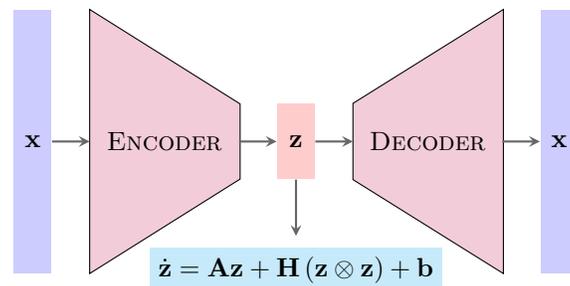

One of major challenges in the approach is to estimate the time-derivative information of $\bz(t)$. If we were to access to the time-derivative of $\bx(t)$, then we can estimate $\dot{\bz}(t)$ as $\tfrac{d\Psi(\bx)}{d\bx} \dot\bx(t)$. But the time-derivative information using time-series data is a challenging task for $\bx$ as well, and even if we manage to have it, computing $\tfrac{d\Psi(\bx)}{d\bx}$ is computationally expensive. As a remedy, we propose to blend a Runge-Kutta  scheme \cite{iserles2009first}---widely used method to integrate differential equations with a high accuracy---to avoid computations or estimate of the derivative information. Yet we obtain a continuous model for the reduced latent variable $\bz(t)$. Here, we focus on the fourth-order Runge-Kutta (\rk) scheme. We briefly recap the \rk~scheme. For a quadratic-nonlinear differential equation \eqref{eq:QuadModel_z}, we can estimate the variable $\bz$ at time $t_{i+1}$ using $\bz$ at time $t_i$ as follows: 
\begin{subequations}
	\begin{align}
		\tilde{\bz}_1 &= \bg(\bz(t_i)) , \quad \tilde{\bz}_2 = \bg\left(\bz(t_i) + \dfrac{h}{2}\tilde{\bz}_1\right), \quad  \tilde{\bz}_3 = \bg\left( \bz(t_i) + \dfrac{h}{2}\tilde{\bz}_3\right), \quad \tilde{\bz}_4 = \bg\left(\bz(t_i) + h \tilde{\bz}_3\right),   \\
		\bz(t_{i+1}) &\approx \bz(t_{i}) + \dfrac{h}{6} \left( \tilde{\bz}_1  + 2\tilde{\bz}_2 + 2\tilde{\bz}_3 + \tilde{\bz}_4\right) =: \Pi_{\text{\rk}}(\bz(t_i)),
	\end{align}
\end{subequations}
where $h = t_{i+1} - t_{i}$ and $\bg(\bz) = \bA\bz + \bH\left(\bz\otimes \bz\right) + \bb$. The \rk~scheme is $\cO(h^4)$ accurate globally, thus for small $h$, estimates are expected to be quite accurate. This allows us to write the problem of identifying quadratic models without requiring derivative information at any stage. 

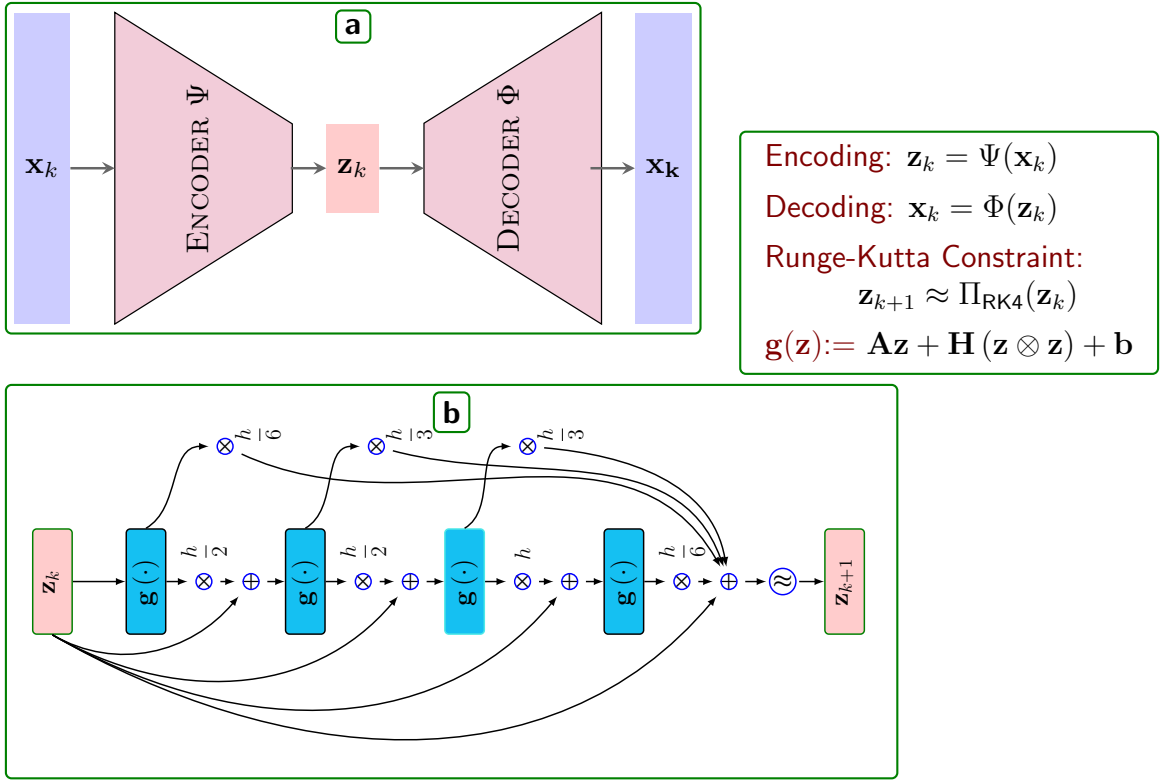
\begin{figure}[tb]
	\tikzset{arrow/.style={-stealth, thick, draw=gray!80!black}}
	\begin{tikzpicture}[font=\sffamily]
		\node[ fill = white,draw = green!50!black, text = black, thick,rounded corners = 0.5ex] (autoencoder) {
			\includegraphics[width= 9.cm]{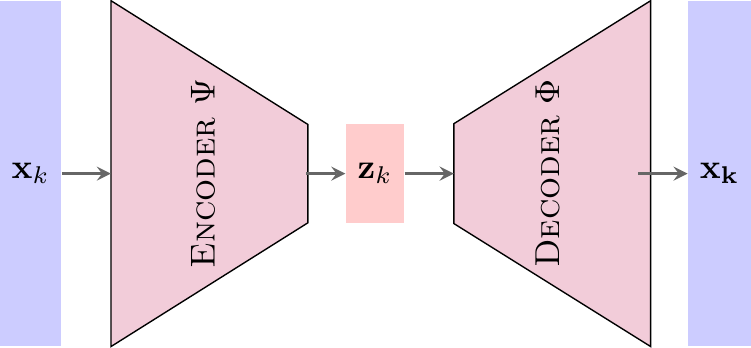}
		};	
		\node[ fill = white,draw = green!50!black, text = black, thick,rounded corners = 0.5ex, below right = -0.5cm and 0.5cm of autoencoder.east, rotate = 0] () { \large 
			\begin{tabular}{l} 
				{\color{red!50!black}Encoding:} $\bz_k = \Psi(\bx_k)$\\[5pt]
				{\color{red!50!black}Decoding:} $\bx_k = \Phi(\bz_k)$\\[5pt]
				{\color{red!50!black}Runge-Kutta Constraint:} \\ ~~~~~~~~~$\bz_{k+1} \approx \Pi_{\text{\rk}}(\bz_k)$\\[5pt]
				{\color{red!50!black}$\bg(\bz)$:=} $\bA\bz + \bH\left(\bz\otimes \bz \right) + \bb$ 
			\end{tabular}
		};	
		%
		\node[ fill = white,draw = green!50!black,   below =  5.5cm of autoencoder.west, text = black, thick,rounded corners = 0.5ex, rotate = 90] (RKsteps) { \includegraphics[width= 5.0cm]{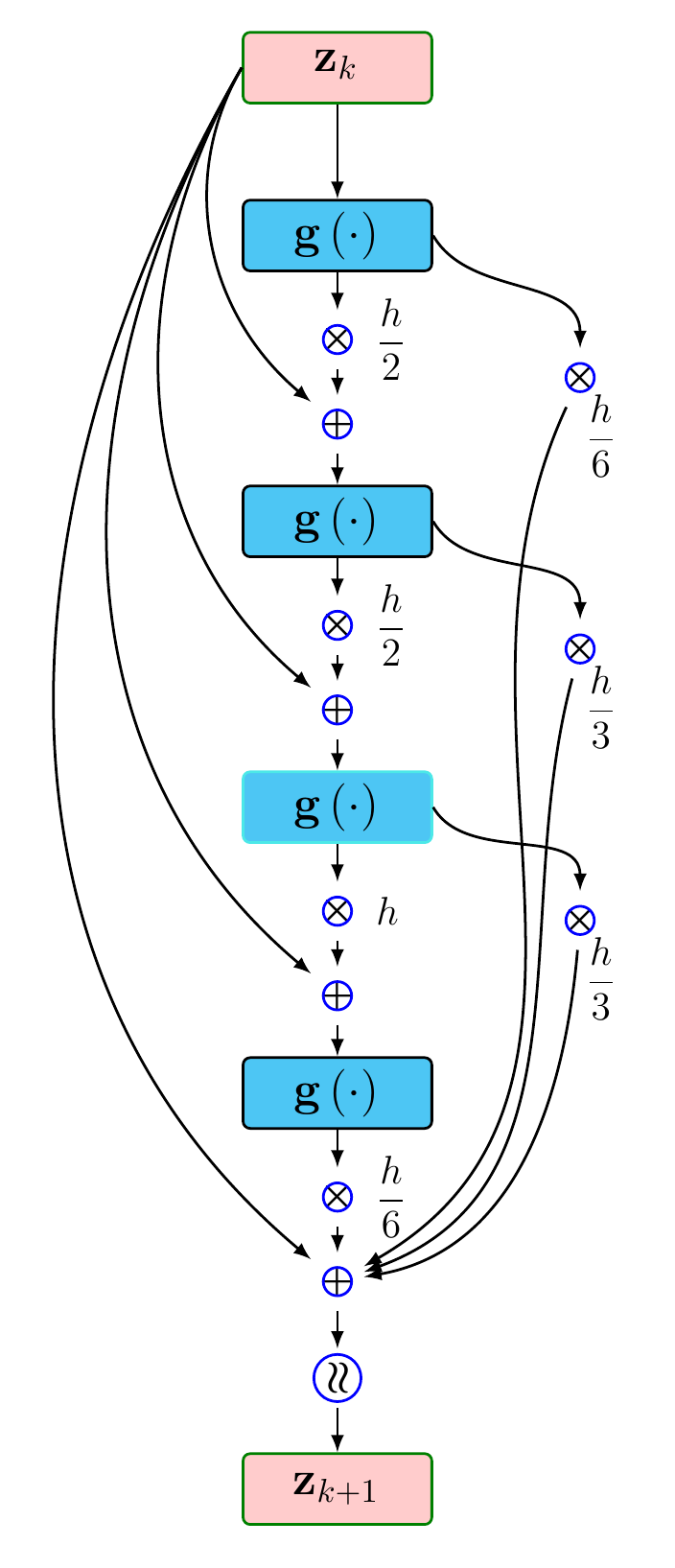}
		};
		\node[ fill = white,draw = green!50!black,   below =  0.1cm of autoencoder.north, text = black, thick,rounded corners = 0.5ex, rotate = 0] () {\large \textbf{a}};
		\node[ fill = white,draw = green!50!black,   below =  0.1cm of RKsteps.east, text = black, thick,rounded corners = 0.5ex, rotate = 0] () {\large \textbf{b}};
		
	\end{tikzpicture}
	\caption{The figure depicts the core principle of the approach to identify low-dimensional quadratic-embeddings of high-fidelity dynamical systems. Our architecture is inspired by autoencoder neural network and a Runge-Kutta scheme. The (\textbf{a}) identifies low-dimensional coordinates $z = \Psi(\bx)$ using encoder and recovers $\bx = \Phi(\bz)$ by decoding it. The (\textbf{b}) shows the fourth-order Runge-Kutta scheme to predict $z_{k+1}$ (the variable $\bz$ at time $t_{k+1}$) from $z_k$  (the variable $\bz$ at time $t_{k}$) so that estimating derivatives can be avoided.}
	\label{fig:core_architecture}
\end{figure}

We depict our core network architecture in \Cref{fig:core_architecture}. As indicated in the figure, we aim at achieving the goals with our architecture, which are: discovering a low-dimensional representation of $\bx$, i.e., $\bz = \Psi(\bx)$, together with dynamics of $\bz$ satisfying $\dot \bz = \bA\bz + \bH\left(\bz\otimes \bz\right) + \bb$. Also, we seek to identify a decoder, mapping $\bz$ to $\bx$, i.e., $\bx = \Phi(\bz)$. For this, we require two types of loss functions to obtain encoder-decoder functions and a quadratic model describing dynamics of the low-dimensional embedding:
\begin{itemize}
	\item \textbf{Reconstruction loss:} First, we seek to determine a low-dimensional embedding $\bz = \Psi(\bx)$, for which we construct a dynamical model, so that the high-fidelity state $\bx$ can again be reconstructed using $\bz$ using an inverse mapping, i.e., $\bx = \Phi(\bz)$. For this, we make use of an autoencoder neural network design as in \Cref{fig:core_architecture}(a). Indeed, the dimension of $\bz$ is a hyper-parameter but the singular values of the data can provide a good indication of it. To training networks or encoder/decoder, we penalize reconstruction inaccuracy by the autoencoder through low-dimensional bottleneck as follows:
	\begin{equation}
		\cL_{\textsf{Rec}} :=	\| \bx - \Phi(\Psi(\bx)) \|,
	\end{equation}
	where $\|\cdot\|$ denotes the mean-squared error, averaging over all the samples and dimensions. 
	\item \textbf{Quadratic dynamics:} Our second goal is to find the low-dimensional variable $\bz$ such that a quadratic model can describe its dynamics, i.e., $\dot \bz(t) = \bA\bz(t) + \bH\left(\bz(t) \otimes \bz(t) \right) + \bb$.  To avoid time-derivative computational or estimates, we enforce the \rk~scheme as illustrated in \Cref{fig:core_architecture}(b). To learn the corresponding quadratic model, we add the following loss function:
	\begin{equation}
		\cL_{\text{\rk}} :=\left\| \bz(t_{i+1}) - \Pi_{\text{\rk}}\left(\bz(t_i)\right) \right\|,
	\end{equation}
	where $\bz(t_{i+1})$ and $\bz(t_i)$ are values at time $t_{i+1}$ and $t_i$, respectively.  Moreover, we can predict its past value at time $t_{i-1}$ using present value of $\bz(t_i)$ by integrating backward in time using the \rk~scheme, i.e., 
	\begin{equation}
		\cL^{\textsf{b}}_{\text{\rk}} :=\left\| \bz(t_{i-1}) - \Pi^{\textsf{back}}_{\text{\rk}}\left(\bz(t_i)\right) \right\|. 
	\end{equation}
	One may add future predictions over $m$-steps using \rk~scheme, but it severely affects training--potentially due to vanishing or exploding gradient issues, and increases computational burden as well. 
\end{itemize}
We combine these losses to train the autoencoder and a quadratic model simultaneously, which is:
\begin{equation}\label{eq:total_loss}
\cL_{\textsf{Total}} := 	\cL_{\textsf{Rec}} + \lambda \left( \cL_{\text{\rk}} + \cL^{\textsf{b}}_{\text{\rk}}  \right),
\end{equation}
where $\lambda$ is a hyper-parameter. 

\begin{remark}
 Here, we have focused on finding low-dimensional coordinate systems for high-fidelity models that can describe the dynamics. Using the proposed methodology, one can also discover an approximate coordinate for low-dimensional nonlinear dynamical systems. The dynamics in the discovered coordinate system can be described as a quadratic model. 
\end{remark}

\section{Demonstration of the Approach}\label{sec:numerics}
We demonstrate the realization of the proposed approach using two examples: nonlinear tabular reactor models and 2D Burgers' equations with a moving shock. We have fixed the parameter $\lambda$ in \eqref{eq:total_loss}  as $\tfrac{1}{\dt}$, where $\dt$ is the time interval between data. 
 In our experiments, we have taken the measurements at a regular interval, though the approach is applicable when measurements are collected at an irregular interval. All experiments were run on A$100$  \nvidia~GPU and have used Pytorch \cite{paszke2019pytorch} to train networks. 
\subsection{Tabular reactor model:} We first consider a 1D tubular reactor model that explains evolution of concentration $\psi(x,t)$ and temperature $\theta(x,t)$. The governing equations are given by partial differential equations \cite{heinemann1981multiplicity}:
\begin{equation}
	\begin{aligned}
		\dfrac{\partial \psi}{\partial t} &= \dfrac{1}{\textsf{Pe}}\dfrac{\partial^2\psi}{\partial x^2} - \dfrac{\partial \psi}{\partial x} - \cD\cF(\psi,\theta;\gamma),\\
		\dfrac{\partial \theta}{\partial t} &=   \dfrac{1}{\textsf{Pe}}\dfrac{\partial^2\theta}{\partial x^2} -\dfrac{\partial \theta}{\partial x} -2.5 \left(\theta-1\right) + 0.5 \cdot \cD\cF(\psi,\theta;\gamma),
	\end{aligned}
\end{equation}
where $x$ being the spatial variable $x \in (0,1)$, time $t>0$, and  Arrhenius reaction term 
\begin{equation}
	\cF(\psi,\theta;\gamma) = \phi \exp\left(\gamma - \dfrac{\gamma}{\theta}\right).
\end{equation}
Moreover, $\cD, \textsf{Pe}$ and $\gamma$ denote Damk\"ohler number, P\'eclet number and the reactor rate, respectively, and we set $D = 0.167, \textsf{Pe} = 5$, and $\gamma = 25$ as discussed e.g., in \cite{zhou2012model,morBenGKetal20}. We omit writing the boundary and initial conditions and refer to \cite{zhou2012model,morBenGKetal20} for them. The model explains an oscillatory dynamics of the reactor.  To collect data, the partial differential equations are discretized using a finite-element scheme by taking $99$ spatial degree of freedom. We gather data at these points  for the concentration and temperature in the time range $t \in [0,60]$ at a regular time interval $\dt = 0.05$. A construction of low-dimensional models of the rector has been considered, e.g., in \cite{zhou2012model,morBenGHetal20} but they require knowledge of either equations or model. 

\begin{figure}[!tb]
	\begin{subfigure}[b]{1\textwidth}
		\centering
		\includegraphics[height = 2.7cm]{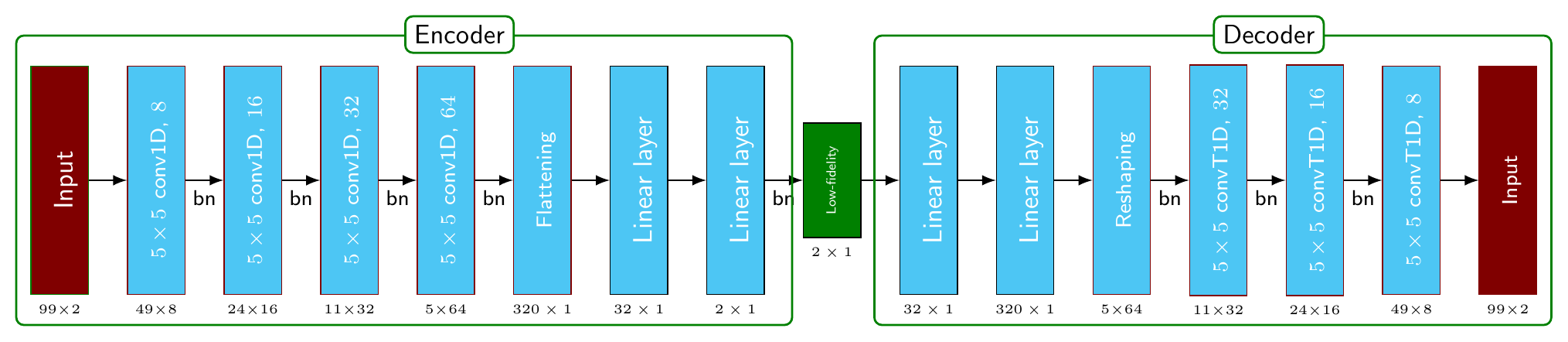}
		\caption{Encoder-decoder design for tabular example.}
		\label{fig:tubular_autoencoder}
	\end{subfigure}
	\begin{subfigure}[b]{1\textwidth}
		\centering
		\includegraphics[height = 2.7cm]{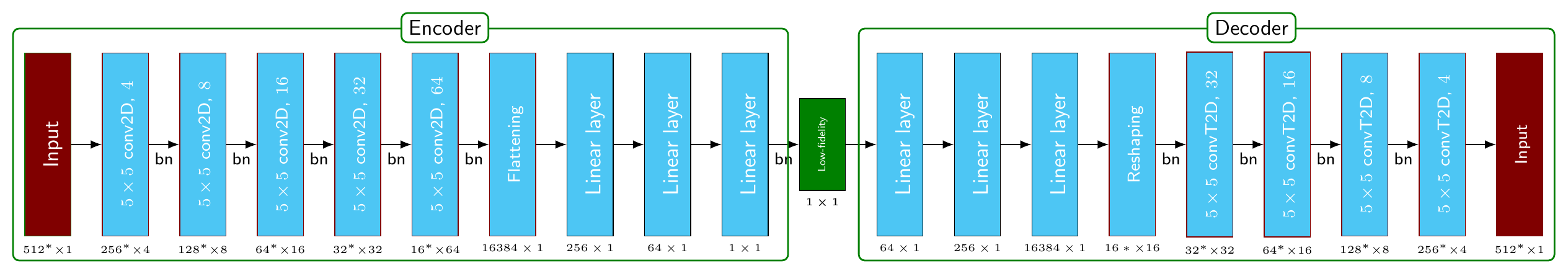}
		\caption{Encoder-decoder design for Burger's example.}
		\label{fig:burgers_autoencoder}
	\end{subfigure}
	\caption{The encoder and decoder architectures are shown in the figure. \emph{\textsf{conv1D}, $k$} (\emph{\textsf{conv2D}, $k$}) indicates a 1D(2D) convolution layer with $k$ kernels of size $5$, and likewise \emph{\textsf{convT1D}, $k$} (\emph{\textsf{convT2D}, $k$}) is a 1D(2D) transpose convolution layer with $k$ transponse kernels of size $5$. We have used \emph{stride} of $2$ to down-sample and up-sample; \textsf{bn} denotes batch-normalization. We have use \textsf{bn} after each convolutional layer and final linear layer in the encoder. We highlight the usage of \textsf{bn} after the last linear layer in the encoder which makes better distribution of the low-dimensional variable, thus improving training. As an activation function, we employed exponential linear unit \cite{clevert2015fast}. Below each block, we denote the size of output of the block.}
\end{figure}
\begin{figure}[!tb]
	\centering
	\begin{subfigure}[b]{0.75\textwidth}
		\centering
		\includegraphics[width=\textwidth]{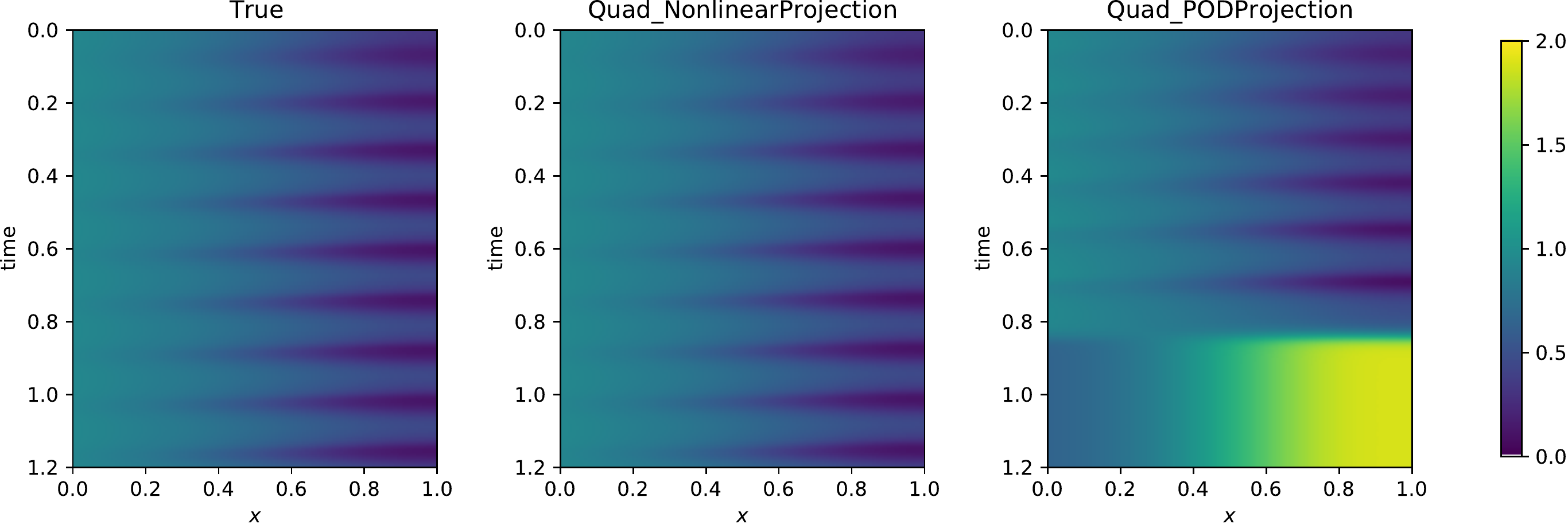}
		\caption{Concentration over time in the domain.}
	\end{subfigure}
	\hfill
	\begin{subfigure}[b]{0.75\textwidth}
		\centering
		\includegraphics[width=\textwidth]{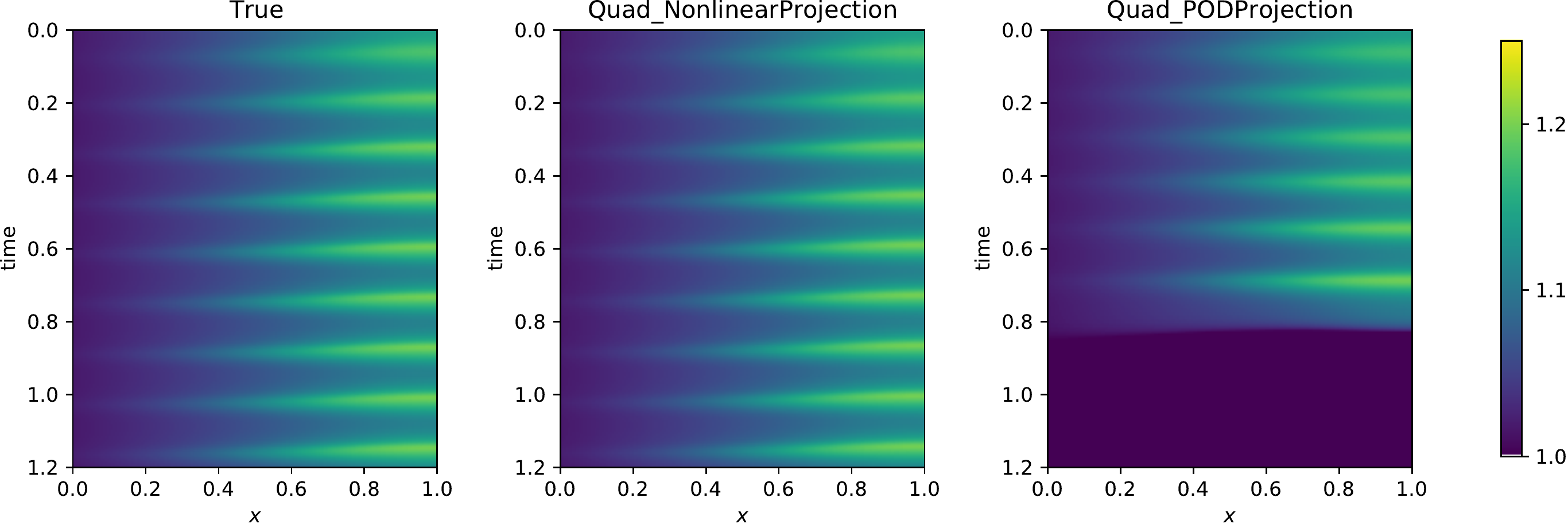}
		\caption{Temperature over time in the domain.}
	\end{subfigure}
	\caption{
		The figure shows a comparison of the accuracy of the encoder-decoder and 2-dimensional quadratic model, which are obtained using the proposed methodology (in the middle) and operator inference in \cite{morPehW16} (in the rightmost). We plot the output of the decoders, which is nothing but the high-fidelity solution, and the input to the decoders is the corresponding 2-dimensional evolution of the embeddings. 
	}
	\label{fig:tubular_sol}
\end{figure}

We construct a low-dimensional model for the reactor using only data. We normalize the data between $0$ and $1$ before training and identifying a low-dimensional quadratic embedding. 
We aim at identifying a low-dimensional model with intrinsic dimension to $\hn = 2$ whose dynamics can be described by a quadratic model. For this, we have used an encoder and decoder architecture design, shown in \Cref{fig:tubular_autoencoder}. We have employed using Adam optimizer \cite{kingma2014adam} for training and have trained for $15~000$ epochs with an initial learning rate $1\cdot 10^{-3}$ which is reduced by one-fifth after every $2~000$ epoch. Once trained encoder, decoder, and the corresponding 2-dimensional quadratic model, we integrate the quadratic model and use the decoder to reconstruct concentration and temperature on the full grid. For comparison, we identify a 2-dimensional coordinate by projecting the full-dimensional solution using the most dominant POD basis and learn the best quadratic model, describing dynamics~\cite{morPehW16}. We plot the reconstruction in \Cref{fig:tubular_sol}, where we observe that the proposed methodology can discover a simple two-dimensional quadratic dynamics model which can describe the dynamics of the high-fidelity system.

\subsection{2D Burgers' equation}
\begin{figure}[tb]
	\includegraphics[width = 0.90\textwidth]{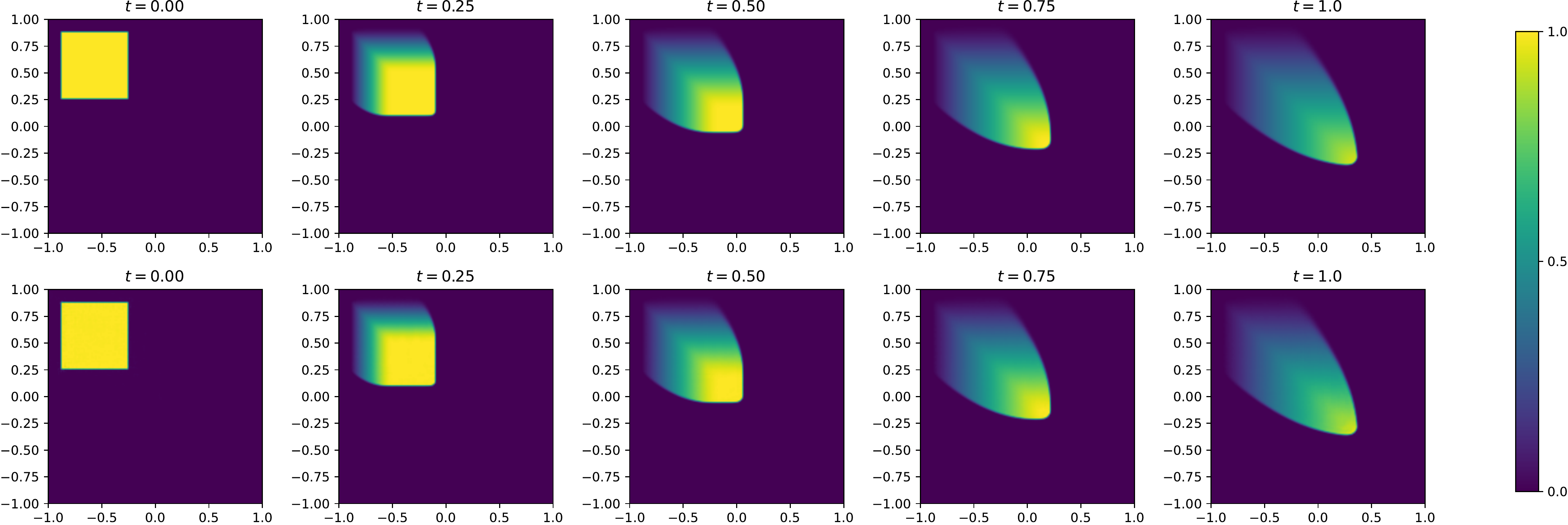}
\caption{
	The transient behavior obtained from the learned models (encoder, decoder, and quadratic model) with the ground truth is shown. The first row shows the ground truth, and the second row is from the learned model, indicating a good low-dimensional surrogate that faithfully captures the original dynamics.}
	\label{fig:burgers}
\end{figure}

In our second example, we consider a  2D Burgers' equation that explains several flow-related dynamics. The dynamics is given by 
\begin{equation}
	\dfrac{\partial u(x,t)}{\partial t} + \left(\dfrac{1}{2}, \dfrac{1}{2} \right)^\top \cdot \nabla u(x,t)^2 = 0, \forall(x,t) \in \Omega \times [0,T].
\end{equation}
We consider a square block as an initial condition as in \cite{sarna2021data}---that is,
\begin{equation}
	u(x,t=0) = \begin{cases}
		1 , \quad x \in [0,0.5]^2\\
		0 , \quad \text{else}
	\end{cases}
\end{equation} 
Like in \cite{sarna2021data}, we also take $\Omega$ to be $(-0.1,1.5)^2$. Having discretized the domain by taken $512$ points in $x$-direction as well as in $y$-direction; thus, the system has $262~144$ degrees of freedom. We take $100$ data in the time interval $[0,1]$ at a regular time grid. We make the encoder and decoder designs as shown in \Cref{fig:burgers_autoencoder} with bottleneck dimension being only \emph{one}. We have trained the networks using Adam for $15~000$ epochs. We set the initial learning rate $1\cdot 10^{-3}$, which is being reduced by one-fifth after each $2~000$ epochs. After trained, we identify a one-dimensional quadratic model that encodes dynamics of complex Burger's equations. We obtain the evolution of the low-dimensional variable by integrating with an appropriate initial condition and reconstruct the solution of the full domain using the decoder. The results are shown in \Cref{fig:burgers}. We mention that we tried to obtain the best one-dimensional quadratic model using linear projection, but the model could not capture any dynamics reasonably; hence, we refrain from plotting them in the figure. It illustrates the impressive capabilities of the neural networks to identify low-dimensional coordinates such that a quadratic model can describe its dynamics. It allows us to come up with low-dimensional using nonlinear projection (encoder-decoder using neural networks) for data with a  slow-decay of  \emph{Kolmogorov n-width}. Hence, for complex high-fidelity nonlinear systems, we can identify a low-dimensional quadratic model, encoding dynamics, thus easing engineering studies, e.g., control, optimization. 


\section{Discussion}\label{sec:conclusion}
In our work, we have utilized the power of deep learning to determine a low-dimensional embedding of high-fidelity dynamical models so that the dynamics of the low-dimensional embedding can be described by a simple quadratic model. To that end, we have employed autoencoder architectures with a constraint of the low-dimensional bottleneck embedding--that is, a quadratic model can explain its dynamics. It is important to learn the low-dimensional representation and the corresponding quadratic model to obtain interpretable and parsimonious models.
The proposed methodology addresses two main drawbacks of earlier existing approaches, which are: 
(a) smooth-enough nonlinear systems can be written as a quadratic model in the proper coordinates, but it is not intuitive to find using only data, and  
(b)  many applications such as advection-dominant problems show a slow decay of the Kolmogorov $n$-width; thus, a low-dimensional representation using a linear projection of data displays a poor performance. In our approach, the usage of an autoencoder and neural networks can be interpreted to find correct low-dimensional representation using a nonlinear projection while ensuring that a quadratic model can explain the dynamics of the representation. We have demonstrated our approach using two examples to determine parsimonious models for complex high-dimensional dynamical systems. 

There are many open avenues for future research. An emerging field of research in science and engineering is \emph{scientific machine learning} (see, e.g., \cite{willcox2021imperative}), in which a primary goal is to infuse empirical knowledge and first-principles rules of processes in the course of training neural networks. We hope that by incorporating such information and physics while training, one learns models using fewer data (otherwise, deep learning approaches are data-demanding). Also, the autoencoder would be better interpretable and generalizable.  Furthermore, finding suitable designs of an encoder and decoder is a concern though there are enough intuitions by deep learning experts to find reasonably good designs. Moreover, we mention that despite our effort to avoid computation of the time-derivatives---which is dubious for noisy measurements---to obtain a continuous quadratic model by fusion with a Runge-Kutta method, our approach still may show poor performance. However, one can execute a de-noising step to remove noise from data using, e.g., the techniques in \cite{rudy2019deep,morGoyB21b}. Last but not least, finding the intrinsic dimension of the low-dimensional embedding with the desired properties plays a crucial role in the performance of our approach. So, to find a good estimate of it, thus leading to the most parsimonious representation, would require further research. 

\section*{Acknowledgment}
We would like to express our gratitude to Dr.\ Neeraj Sarna for providing data for 2D Burgers' example considered in Subsection 4.2. 
\section*{Funding Statement}
Peter Benner was partially supported by the German Research Foundation (DFG) Research Training Group 2297 ``MathCoRe'', Magdeburg. 

\addcontentsline{toc}{section}{References}
\bibliographystyle{ieeetr}
\bibliography{mybib}

\end{document}